\documentclass[conference]{IEEEtran}
\IEEEoverridecommandlockouts
\usepackage{cite}
\usepackage{amsmath,amssymb,amsfonts}
\usepackage{algorithmic}
\usepackage{graphicx}
\usepackage{textcomp}
\usepackage{xcolor}
\def\BibTeX{{\rm B\kern-.05em{\sc i\kern-.025em b}\kern-.08em
    T\kern-.1667em\lower.7ex\hbox{E}\kern-.125emX}}

\begin{document}

\title{A Comparative Study of Machine Learning Algorithms for Electricity Price Forecasting with LIME-Based Interpretability}

\author{
\begin{minipage}{\textwidth}
\centering
\begin{tabular}{cc}

% Top-left author
\begin{tabular}[t]{@{}c@{}}
Xuanyi Zhao \\
\textit{Business and Tourism School} \\
\textit{Sichuan Agricultural University} \\
Chengdu, China \\
xuanyizhao912@gmail.com
\end{tabular}
&
% Top-right author
\begin{tabular}[t]{@{}c@{}}
Jiawen Ding \\
\textit{School of Computing and Information Systems} \\
\textit{The University of Melbourne} \\
Melbourne, Australia \\
jiawending@student.unimelb.edu.au
\end{tabular}

\\[3em]

% Bottom-left author
\begin{tabular}[t]{@{}c@{}}
Xueting Huang \\
\textit{School of Science, Computing and Engineering Technologies} \\
\textit{Swinburne University of Technology} \\
Melbourne, Australia \\
xuetinghuang@swin.edu.au
\end{tabular}
&
% Bottom-right author
\begin{tabular}[t]{@{}c@{}}
Yibo Zhang \\
\textit{Gezhi Future Research Institute} \\
Beijing, China \\
bernie.zhangyibo@gmail.com
\end{tabular}

\end{tabular}
\end{minipage}
}

\maketitle

\begin{abstract}
With the rapid development of electricity markets, price volatility has significantly increased, making accurate forecasting crucial for power system operations and market decisions. Traditional linear models cannot capture the complex nonlinear characteristics of electricity pricing, necessitating advanced machine learning approaches. This study compares eight machine learning models using Spanish electricity market data, integrating consumption, generation, and meteorological variables. The models evaluated include linear regression, ridge regression, decision tree, KNN, random forest, gradient boosting, SVR, and XGBoost. Results show that KNN achieves the best performance with R² of 0.865, MAE of 3.556, and RMSE of 5.240. To enhance interpretability, LIME analysis reveals that meteorological factors and supply-demand indicators significantly influence price fluctuations through nonlinear relationships. This work demonstrates the effectiveness of machine learning models in electricity price forecasting while improving decision transparency through interpretability analysis.
\end{abstract}

\begin{IEEEkeywords}
electricity price forecasting, machine learning, K-nearest neighbors, LIME, interpretability analysis
\end{IEEEkeywords}

\section{Introduction}

With the rapid evolution of global energy structures and the gradual liberalization of electricity markets, electricity price volatility has significantly increased. Accurate electricity price forecasting has become a critical issue in power system operations, market transactions, and policy-making. Influenced by factors such as the rising share of renewable energy, increased demand-side uncertainty, and climate change, traditional electricity dispatching and market mechanisms are facing unprecedented challenges\cite{b1}. Therefore, developing price forecasting models with high accuracy and strong generalization capabilities is essential not only for optimizing the allocation of electricity resources but also for enhancing the economic efficiency and stability of market operations\cite{b2}. In recent years, machine learning techniques have demonstrated remarkable advantages in time series modeling and complex relationship mining, offering new research directions and technical pathways for electricity price forecasting\cite{b3}. Electricity price forecasting is not only a theoretical problem but also a critical tool for practical applications. For example, accurate forecasts assist electricity providers in optimizing power generation scheduling and reducing operational risks. In smart grid systems, timely price signals are essential for enabling demand response and integrating distributed energy resources such as solar and wind. Industrial consumers rely on price predictions to adjust their energy procurement strategies, while market regulators use them to ensure fair pricing and grid stability\cite{b4}.

Although various studies have explored the application of different models to electricity price forecasting, several limitations remain: (1) the selection of models often lacks systematic comparison, making it difficult to comprehensively assess the applicability and performance differences of various algorithms\cite{b4}; (2) most research focuses on forecasting accuracy while neglecting model interpretability, thereby limiting its practical value in real market environments\cite{b5}; (3) the influence mechanisms of key features on prediction outcomes are insufficiently revealed, hindering the provision of transparent and reliable decision-making support to market participants\cite{b6}.

To address these issues, this paper proposes an electricity price forecasting approach based on multiple machine learning algorithms and introduces the LIME interpretability framework to explore model performance, result interpretability, and key factor identification in depth. The main contributions of this study are as follows:

\begin{itemize}
    \item A systematic comparison of eight mainstream machine learning models in electricity price forecasting tasks\cite{b7};
    \item The application of the LIME method to conduct local interpretability analysis of the optimal model, thereby improving model transparency\cite{b8};
    \item Identification and quantification of key features that influence electricity price fluctuations, and the elucidation of their nonlinear interaction mechanisms\cite{b9}.
\end{itemize}

The structure of this paper is organized as follows: Section 2 presents the dataset used, preprocessing procedures, and selected modeling methods. Section 3 reports the evaluation results of model performance and analyzes the visualization outputs and LIME-based explanations of the optimal model. Section 4 summarizes the key findings and discusses the practical implications and future directions of this research.

\section{Methods}
\subsection{Data Sources and Preprocessing}

This study uses a publicly available dataset from Kaggle, which contains four years of electricity consumption, generation, price, and weather data in Spain. The dataset includes hourly records of electricity demand, generation, and settlement prices in the Spanish electricity market, along with weather information from five major cities in Spain. Electricity consumption and generation data are obtained from the European Network of Transmission System Operators for Electricity (ENTSOE), settlement prices are sourced from Red Eléctrica de España, and weather data are purchased from the Open Weather API\cite{b10}.

The raw dataset consists of multiple files\cite{b11}. The energy data includes variables such as real-time generation, generation forecasts, actual load, and load forecasts, categorized by energy type (e.g., nuclear, wind, solar)\cite{b12}. Weather data covers meteorological parameters such as temperature, humidity, precipitation, and wind speed\cite{b13}. All data are recorded on an hourly basis from 2015 to 2018, totaling approximately 35,000 data points. Due to the diverse sources, the dataset contains issues such as time misalignment, inconsistent formats, and missing values, necessitating a systematic preprocessing procedure\cite{b14}.

\begin{figure}[htbp]
\centering
\includegraphics[width=0.5\textwidth]{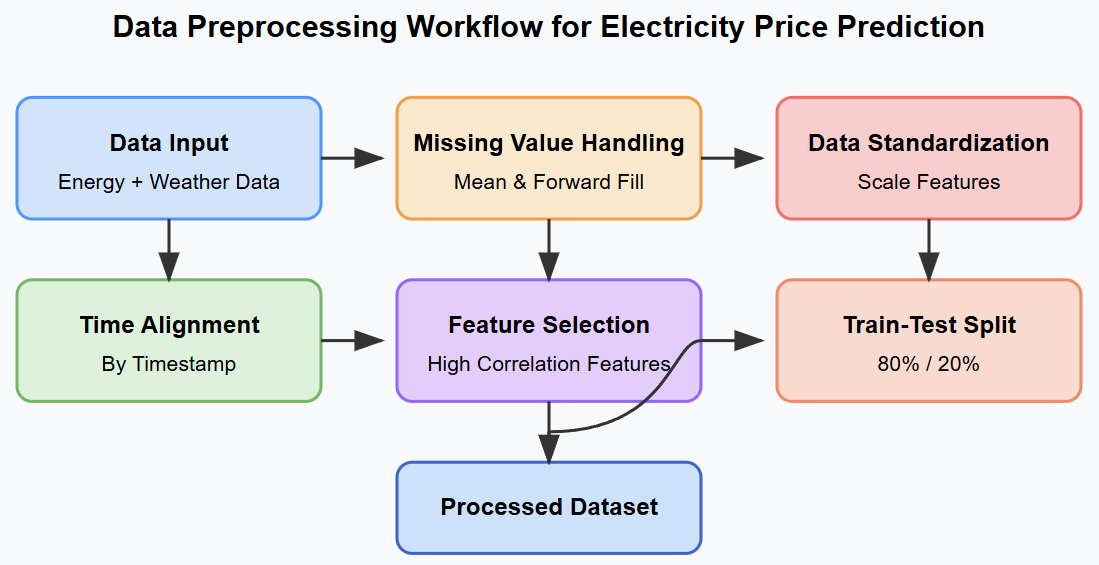}
\caption{Data Preprocessing Flow for Electricity Price Prediction}
\label{fig:preprocessing}
\end{figure}

The data preprocessing pipeline, illustrated in Figure~\ref{fig:preprocessing}, includes the following main steps:

\begin{enumerate}
  \item \textbf{Data Integration}: Energy data (consumption and generation) and weather data are merged. Generation data from different energy types are first consolidated into a unified structure and then merged with weather data using timestamp alignment via the \texttt{merge} function in \texttt{pandas}, ensuring all data are analyzed under a unified framework.
  
  \item \textbf{Time Alignment}: Data from different sources are aligned based on timestamps. All timestamps are converted to a unified UTC format, and new temporal features such as hour, date, and month are created to capture temporal patterns in electricity pricing.
  
  \item \textbf{Missing Value Handling}: Missing values are addressed using a combination of mean imputation and forward filling. For short-term missing weather data, linear interpolation is applied. For missing energy data, imputation strategies are selected based on the variable characteristics, such as using values from previous time steps or type-specific averages.
  
  \item \textbf{Feature Selection}: Features highly correlated with electricity prices are selected. Pearson correlation analysis and feature importance evaluation are conducted to identify variables that significantly contribute to price forecasting. A total of 27 core features are retained, including various energy outputs, meteorological indicators, and time features.
  
  \item \textbf{Data Standardization}: Features are standardized to bring all values into a comparable range. The \texttt{StandardScaler} is used to scale numerical features to have zero mean and unit variance, ensuring balanced contributions of features to model training.
\end{enumerate}

After the above steps, the raw data are transformed into a structured feature matrix, providing high-quality inputs for subsequent model training and evaluation. The processed dataset contains approximately 33,500 valid records, preserving the temporal characteristics of the original data while eliminating outliers and inconsistencies.

\subsection{Model Construction}

This study builds eight machine learning models for electricity price forecasting, covering linear models, tree-based models, instance-based models, and ensemble learning methods, aiming to comprehensively compare the performance of different algorithm types in this task. Model selection balances computational efficiency, predictive accuracy, and interpretability to meet practical application needs\cite{b15}.

Linear regression serves as the baseline model, which minimizes mean squared error to identify linear relationships between features and electricity prices. Although simple and computationally efficient, it struggles to capture complex nonlinear relationships. Ridge regression extends linear regression by introducing an L2 regularization term, effectively mitigating multicollinearity and overfitting, thereby improving generalization on unseen data. In this study, the regularization parameter $\alpha$ is determined to be 0.1 via cross-validation.

The decision tree model uses the Gini index as the splitting criterion, with a maximum depth of 8 to prevent overfitting. It can automatically discover nonlinear relationships and interactions between features and offers good interpretability. The k-nearest neighbors (KNN) algorithm predicts the electricity price of a new sample based on a distance metric in feature space, using a weighted average of the $k$ nearest neighbors’ prices. In this study, the number of neighbors $k$ is set to 5 via grid search, Euclidean distance is used as the metric, and weights are inversely proportional to distance.

Random forest, as an ensemble of decision trees, trains multiple independent trees and aggregates their predictions to reduce the variance of individual trees, thereby improving model stability and accuracy. In this experiment, the number of trees is set to 100, and the feature subset sampling ratio is 0.7. Gradient boosting trees sequentially train a set of trees, where each new tree focuses on reducing the residuals from the previous stage, enabling powerful predictive performance. The learning rate is set to 0.1, and the maximum tree depth is 4 to control model complexity.

Support vector regression (SVR) uses the kernel trick to map data to a high-dimensional feature space and seeks the maximum-margin hyperplane for regression in that space. This study uses the radial basis function (RBF) kernel, with kernel parameter $\gamma = 0.1$ and regularization parameter $C = 1.0$. XGBoost, a highly efficient implementation of gradient boosting, incorporates regularization and parallel computing capabilities, performing exceptionally well on large-scale datasets. Key parameters for XGBoost in this study include: learning rate of 0.05, maximum depth of 6, subsample rate of 0.8, and feature sampling rate of 0.7.

All models are trained on the same training set, with hyperparameters optimized using grid search and 5-fold cross-validation. To ensure a fair comparison, all models are evaluated on the same test set using consistent performance metrics. Model implementation is based on the \texttt{scikit-learn} and \texttt{XGBoost} libraries, and training is conducted on a computing environment with 16GB RAM and a quad-core CPU.

\subsection{Model Evaluation}

To comprehensively evaluate model performance, this study adopts six evaluation metrics: MAE, MSE, RMSE, $R^2$, MAPE, and EVS. These indicators reflect the prediction error, relative deviation, and model fitting ability from multiple perspectives. The definitions and mathematical formulas of each metric are as follows:

\begin{enumerate}
    \item \textbf{MAE (Mean Absolute Error)}: Measures the average absolute difference between predicted and actual values:
    \[
    \text{MAE} = \frac{1}{n} \sum_{i=1}^{n} \left| y_i - \hat{y}_i \right|
    \]
    
    \item \textbf{MSE (Mean Squared Error)}: Measures the average squared difference between predicted and actual values:
    \[
    \text{MSE} = \frac{1}{n} \sum_{i=1}^{n} (y_i - \hat{y}_i)^2
    \]
    
    \item \textbf{RMSE (Root Mean Squared Error)}: The square root of MSE, sensitive to large deviations:
    \[
    \text{RMSE} = \sqrt{ \frac{1}{n} \sum_{i=1}^{n} (y_i - \hat{y}_i)^2 } = \sqrt{\text{MSE}}
    \]
    
    \item \textbf{$R^2$ (Coefficient of Determination)}: Indicates how well the model explains the variance in the actual data:
    \[
    R^2 = 1 - \frac{\sum_{i=1}^{n}(y_i - \hat{y}_i)^2}{\sum_{i=1}^{n}(y_i - \bar{y})^2}
    \]
    
    \item \textbf{MAPE (Mean Absolute Percentage Error)}: Measures the average relative error:
    \[
    \text{MAPE} = \frac{1}{n} \sum_{i=1}^{n} \left| \frac{y_i - \hat{y}_i}{y_i} \right| \times 100\%
    \]
    
    \item \textbf{EVS (Explained Variance Score)}: Measures the proportion of variance explained by the model:
    \[
    \text{EVS} = 1 - \frac{\text{Var}(y - \hat{y})}{\text{Var}(y)}
    \]
\end{enumerate}

In this study, MAE, MSE, and RMSE are used to reflect absolute error levels; $R^2$ and EVS represent the model’s ability to explain data variability; MAPE serves as a supplementary metric to evaluate relative prediction performance. Together, these metrics provide a comprehensive and objective basis for model evaluation.

\subsection{LIME Interpretability Analysis}

To interpret the prediction results of the model, this study employs the LIME (Local Interpretable Model-agnostic Explanations) method for local interpretability analysis. LIME constructs a locally linear surrogate model near the target prediction instance to reveal the local decision boundary of the black-box model and enhance transparency in prediction.

The mathematical expression of LIME is as follows:
\[
\xi(x) = \arg\min_{g \in G} \mathcal{L}(f, g, \pi_x) + \Omega(g)
\]
where \( \xi(x) \) represents the explanation for instance \( x \), \( f \) is the original complex model, \( g \) is the local surrogate model, \( \pi_x \) is the proximity weighting function around \( x \), \( \mathcal{L} \) is the local fidelity loss function, and \( \Omega(g) \) denotes the complexity penalty of the surrogate model.

The implementation of LIME includes the following steps:

\begin{enumerate}
    \item \textbf{Sampling}: Generate a set of perturbation samples \( \{z_1, z_2, \dots, z_n\} \) around the target instance \( x \), where the perturbation range is constrained to ensure the standard deviation of features remains within a certain ratio.
    
    \item \textbf{Weighting}: Compute the similarity between each perturbed sample and the original instance \( x \), using a Gaussian kernel as the weight function:
    \[
    \pi_x(z) = \exp\left(-\frac{D(x, z)^2}{\sigma^2}\right)
    \]
    where \( D(x, z) \) is the Euclidean distance, and \( \sigma \) is the kernel width.
    
    \item \textbf{Prediction}: Use the original model \( f \) to predict the outcomes of all perturbed samples, obtaining \( \{f(z_1), f(z_2), \dots, f(z_n)\} \).
    
    \item \textbf{Fitting a Surrogate Model}: Fit a local surrogate model \( g \) on the perturbed dataset using the weighted loss function:
    \[
    \mathcal{L}(f, g, \pi_x) = \sum_{i=1}^{n} \pi_x(z_i) \cdot (f(z_i) - g(z_i))^2
    \]
    
    \item \textbf{Feature Contribution Analysis}: Analyze the coefficients of the surrogate model to quantify the contribution of each feature to the prediction result.
\end{enumerate}

In this study, the number of perturbation samples is set to 300. The Euclidean distance is used for similarity calculation, and L2 regularization is applied to control model complexity. LIME analysis results demonstrate that meteorological features and supply-demand indicators significantly influence electricity price prediction, offering insights into the underlying decision logic of the model.

\section{Experimental Results and Analysis}
\subsection{Model Performance Comparison}

This study implements and evaluates eight machine learning models for electricity price forecasting. Table~\ref{tab:performance} shows significant performance differences across models, reflecting the complexity of electricity price data and the challenges of this task.

Linear regression and ridge regression, as baseline models, yield nearly identical results with $R^2 = 0.432$, MAE around 8.2, and RMSE exceeding 10.7. This suggests that linear models struggle to capture the complex relationships in the data. The decision tree improves upon this with $R^2 = 0.642$ and RMSE reduced to 8.525, indicating its ability to handle nonlinearity to some extent.

\begin{table}[htbp]
\caption{Performance Evaluation of Different Models}
\label{tab:performance}
\centering
\begin{tabular}{lcccccc}
\hline
\textbf{Model} & \textbf{MAE} & \textbf{MSE} & \textbf{RMSE} & \textbf{$R^2$} & \textbf{MAPE} & \textbf{EVS} \\
\hline
Linear            & 8.199 & 115.377 & 10.741 & 0.432 & 0.172 & 0.432 \\
Ridge             & 8.199 & 115.376 & 10.741 & 0.432 & 0.172 & 0.432 \\
Decision Tree     & 5.154 & 72.677  & 8.525  & 0.642 & 0.101 & 0.643 \\
KNN               & 3.556 & 27.459  & 5.240  & 0.865 & 0.069 & 0.865 \\
Random Forest     & 3.616 & 28.606  & 5.348  & 0.859 & 0.074 & 0.859 \\
GradBoost         & 6.707 & 79.210  & 8.900  & 0.610 & 0.140 & 0.610 \\
SVR               & 5.963 & 72.514  & 8.516  & 0.643 & 0.126 & 0.643 \\
XGBoost           & 4.272 & 33.401  & 5.779  & 0.836 & 0.085 & 0.836 \\
\hline
\end{tabular}
\end{table}

Support vector regression (SVR) and gradient boosting show slight improvements over decision trees, but gradient boosting has a relatively high MAPE of 0.14. XGBoost achieves better performance with $R^2 = 0.836$ and RMSE = 5.779, highlighting its effectiveness.

Among all models, the k-nearest neighbors (KNN) model performs best with $R^2 = 0.865$, MAE = 3.556, RMSE = 5.240, and MAPE = 0.069, indicating excellent predictive accuracy. Random forest follows closely with $R^2 = 0.859$ and similar error metrics.

The results in Table~\ref{tab:performance} show that nonlinear models outperform linear ones in electricity price prediction. Instance-based (KNN) and tree-based ensemble methods (Random Forest, XGBoost) deliver the best results, confirming the presence of nonlinear patterns and local dependencies in electricity price data.

Based on its outstanding performance, the KNN model is selected for subsequent LIME analysis to explore the key factors influencing electricity price fluctuations.

\subsection{Visualization of Prediction Results}

\begin{figure}[htbp]
\centerline{\includegraphics[width=0.4\textwidth]{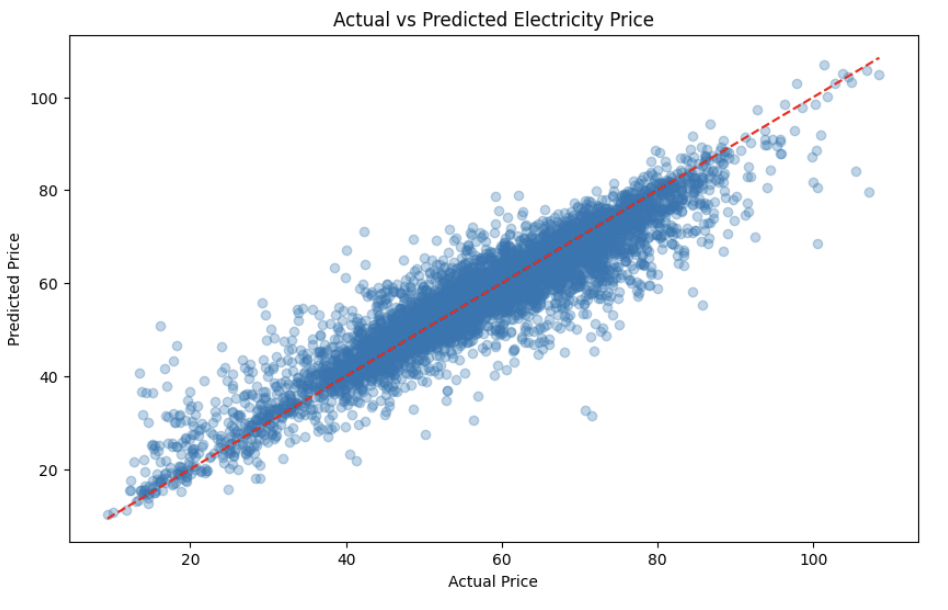}}
\caption{Scatter plot of actual vs. predicted electricity prices.}
\label{fig:scatter}
\end{figure}

Figure~\ref{fig:scatter} shows the scatter plot of actual electricity prices versus KNN model predictions. Most points are concentrated around the diagonal line, indicating strong consistency between the predicted and actual values. However, as the price increases, the dispersion of prediction points becomes more noticeable, suggesting that the model's accuracy slightly decreases in high-price regions.

\begin{figure}[htbp]
\centerline{\includegraphics[width=0.7\linewidth,height = 0.15\textwidth]{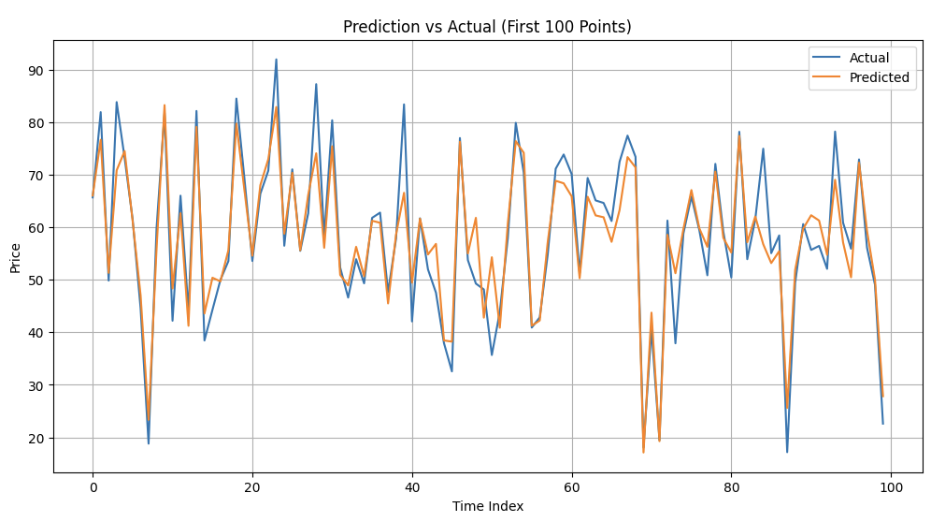}}
\caption{Time series comparison of actual and predicted prices (first 100 steps).}
\label{fig:tsplot}
\end{figure}

Figure~\ref{fig:tsplot} presents the time series comparison of actual and predicted electricity prices over the first 100 time steps. As shown, the KNN model captures the overall trend and fluctuation patterns of electricity prices well. However, certain discrepancies remain at turning points where prices change sharply.

\begin{figure}[htbp]
\centerline{\includegraphics[width=0.6\linewidth,height = 0.15\textwidth]{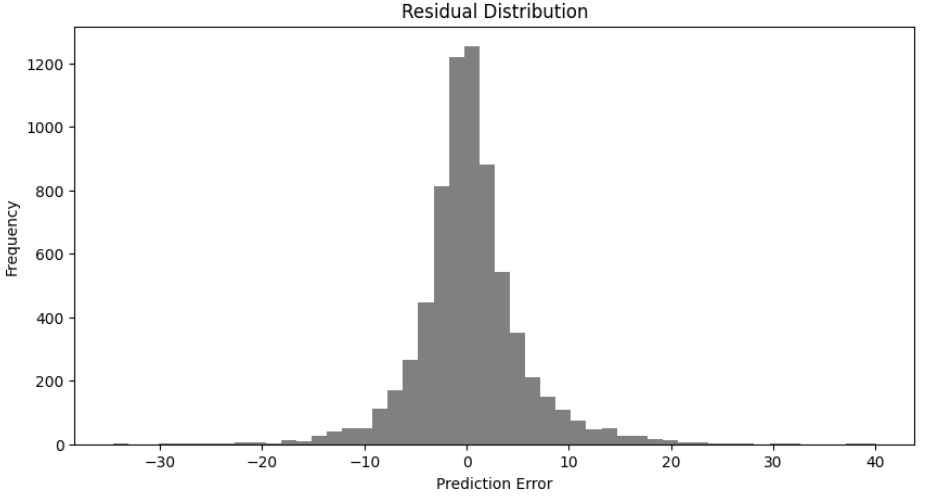}}
\caption{Distribution of prediction errors.}
\label{fig:error}
\end{figure}

Figure~\ref{fig:error} illustrates the distribution of prediction errors for the KNN model. The error distribution is approximately normal, with most values falling within the range of -10 to 10. This indicates that the model’s predictions are generally unbiased, showing no significant tendency to systematically overestimate or underestimate electricity prices.

\subsection{LIME Interpretability Analysis}

\begin{figure}[htbp]
\centerline{\includegraphics[width=0.65\linewidth,height = 0.25\textwidth]{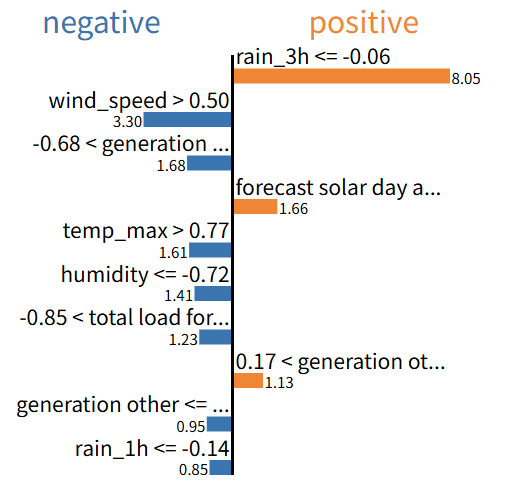}}
\caption{Feature contributions visualized by LIME.}
\label{fig:lime}
\end{figure}

To better understand the KNN model's decision logic, we used the LIME method to conduct local interpretability analysis. Figure~\ref{fig:lime} presents a typical LIME output, showing how different features contribute positively or negatively to a single prediction.

As shown in the figure, meteorological features dominate. For instance, low rainfall (rain\_3h $\leq -0.06$) strongly increases predicted prices, likely due to reduced hydropower availability. Conversely, high wind speed (wind\_speed $> 0.50$) lowers prices by boosting wind energy supply. Other important contributors include solar forecasts, temperature, humidity, and load predictions, each affecting price through supply-demand interactions.

In summary, LIME highlights three insights: 
\begin{itemize}
    \item Weather factors, especially rainfall and wind speed, are key drivers;
    \item Feature effects are nonlinear and threshold-based;
    \item Supply-demand variables jointly shape price outcomes.
\end{itemize}

These findings enhance model transparency and practical interpretability.

\section{Conclusion}

Based on the experimental results, the K-Nearest Neighbors (KNN) model demonstrates superior performance in electricity price forecasting. It effectively captures nonlinear characteristics and local patterns in price data, outperforming linear models and other ensemble learning methods. Furthermore, using the Local Interpretable Model-Agnostic Explanations (LIME) method, we conducted interpretability analysis to identify the key drivers of price fluctuations. The results highlight meteorological variables—such as rainfall and wind speed—as dominant contributors, while power generation and demand-related features also play important roles.

This analysis not only confirms the predictive capability of the KNN model but also enhances its transparency and credibility in real-world applications. In summary, the integrated forecasting and interpretation framework proposed in this study provides a reliable solution that balances both accuracy and interpretability for electricity price forecasting, offering valuable data-driven support for understanding price formation mechanisms and aiding decision-makers in the energy market.

\end{document}